\DocumentMetadata{%
	pdfstandard=A-3u,
	pdfversion=1.7,
	lang=en-US,
}

\documentclass[nocopyright,upint,varvw,barcolor=Goldenrod3,mathalfa=cal=euler,balance,hyphenate,german]{asmejour}

\usepackage{enumitem}
\usepackage{soul} 

\hypersetup{%
	pdfauthor={Chen-Lung Lu, Honglu He, Jinhan Ren, Joni Dhar, Glenn Saunders, Agung Julius, Johnson Samuel, John T.~Wen},                       		   	
	pdftitle={Multi-Robot Scan-n-Print for Wire Arc Additive Manufacturing},                  	
	pdfkeywords={WAAM, Welding Robot, Multiple Robot,  Scan-n-Plan, Welding Process Control},
	pdfsubject = {Describes the proposed scan-n-print framework in the multi-robot scenario},			
}

\JourName{ASME Letters in Translational Robotics}

\newif\ifresubmissionred
\resubmissionredfalse

\newif\ifmodification
\modificationfalse

\begin{document}


\SetAuthorBlock{Chen-Lung Lu\CorrespondingAuthor}
   {Department of Electrical, Computer, and Systems Engineering,\\
   Rensselaer Polytechnic Institute,\\
   Troy, NY 12180 USA \\
   email: luc5@rpi.edu}
\SetAuthorBlock{Honglu He}
   {Department of Electrical, Computer, and Systems Engineering,\\
   Rensselaer Polytechnic Institute,\\
   Troy, NY 12180 USA \\
   email: heh6@rpi.edu}
\SetAuthorBlock{Jinhan Ren}
   {Department of Mechanical, Aerospace, and Nuclear Engineering,\\
   Rensselaer Polytechnic Institute,\\
   Troy, NY 12180 USA \\
   email: renj2@rpi.edu}
\SetAuthorBlock{Joni Dhar}
   {Department of Mechanical, Aerospace, and Nuclear Engineering,\\
   Rensselaer Polytechnic Institute,\\
   Troy, NY 12180 USA \\
   email: dharj@rpi.edu}
\SetAuthorBlock{Glenn Saunders}
   {
   Manufacturing Innovation Center \\
   Rensselaer Polytechnic Institute,\\
   Troy, NY 12180 USA \\
   email: saundg@rpi.edu}
\SetAuthorBlock{Agung Julius}
   {
   Department of Electrical, Computer, and Systems Engineering,\\
   Rensselaer Polytechnic Institute,\\
   Troy, NY 12180 USA \\
   email: agung@ecse.rpi.edu\\ juliua2@rpi.edu}
\SetAuthorBlock{Johnson Samuel}
   {
   Department of Mechanical, Aerospace, and Nuclear Engineering,\\
   Rensselaer Polytechnic Institute,\\
   Troy, NY 12180 USA \\
   email: samuej2@rpi.edu}
\SetAuthorBlock{John T.~Wen}
   {
   Department of Electrical, Computer, and Systems Engineering,\\
   Rensselaer Polytechnic Institute,\\
   Troy, NY 12180 USA \\
   email: wenj@rpi.edu}


\title{Multi-Robot Scan-n-Print for Wire Arc Additive Manufacturing}


\begin{abstract}
Robotic Wire Arc Additive Manufacturing (WAAM) is a metal additive manufacturing technology 
offering flexible 3D printing while ensuring high-quality near-net-shape final parts. However, WAAM also suffers from geometric imprecision, especially for low-melting-point metal such as aluminum alloys.
In this paper, we present a multi-robot framework for WAAM process monitoring and control.  We consider a three-robot setup: a 6-dof welding robot, a 2-dof trunnion platform, and a 6-dof sensing robot with a wrist-mounted laser line scanner measuring the printed part height profile.  
The welding parameters, including the wire feed rate, are held constant based on the materials used, so the control input is the robot path speed.  The measured output is the part height profile.  The planning phase decomposes the target shape into slices of uniform height. 
During runtime, the sensing robot scans each printed layer, and the robot path speed for the next layer is adjusted based on the deviation from the desired profile. The adjustment is based on an identified model correlating the path speed to changes in height.  The control architecture coordinates the synchronous motion and data acquisition between all robots and sensors.  
Using a three-robot WAAM testbed, we demonstrate significant improvements of the closed loop scan-n-print approach over the current open loop result on both a flat wall and a more complex turbine blade shape. 
\end{abstract}

\date{Version \versionno, \today}

\maketitle 


\section{INTRODUCTION}
\label{sec:intro}

Wire Arc Additive Manufacturing (WAAM) is a metal additive manufacturing technology that builds layers of a target shape by melting metal wire using an electric arc, often in conjunction with a motion system \cite{Williams2016wires}. 
By combining the welding system with an industry robot, a robotic WAAM system can produce large format near-net-shape  parts at high deposition rate.  It is ideally suited for rapid metal prototyping and repairs of forged parts in aerospace and automotive industries \cite{Busachi2015designing}. 

\begin{figure}[ht]
    \centering    \includegraphics[width=0.47\textwidth]{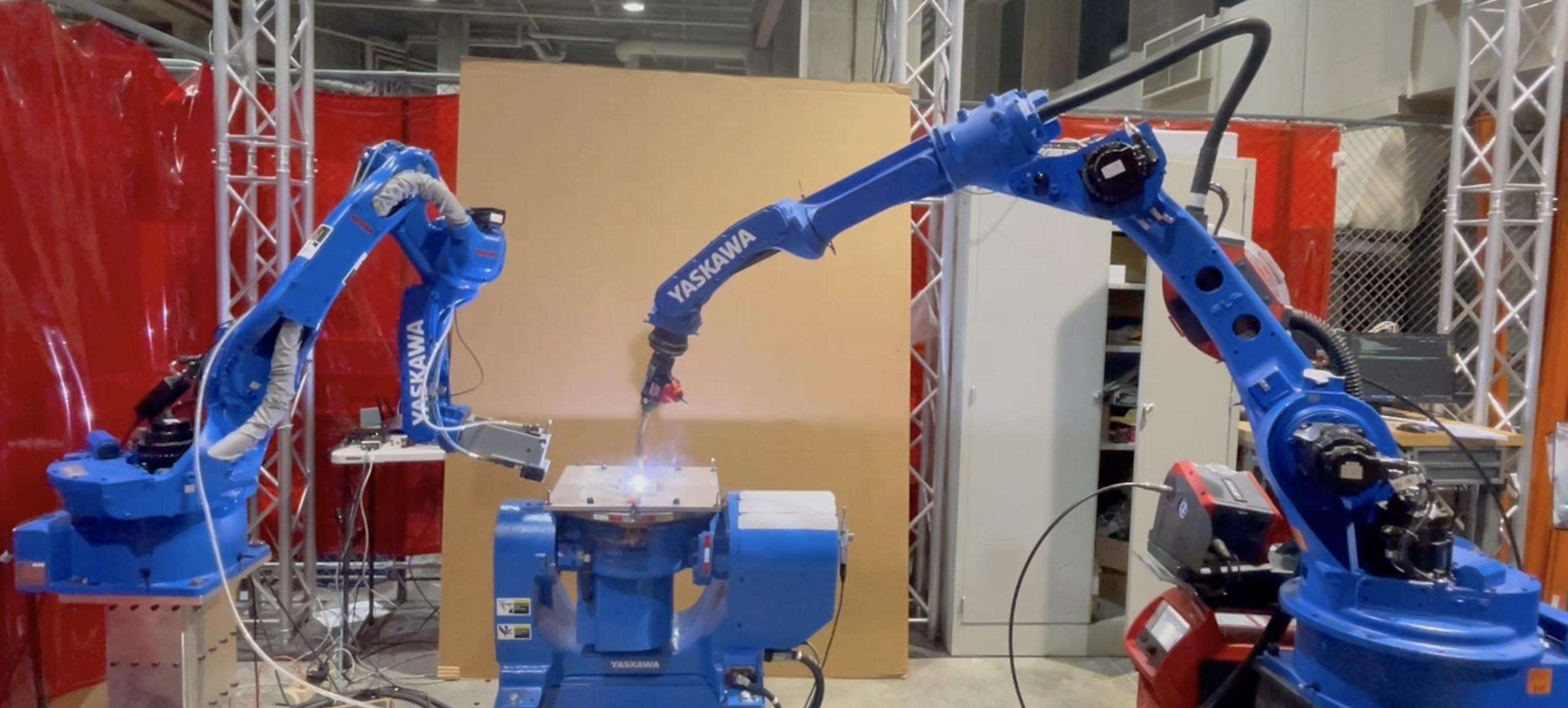}
    \caption{Scan-n-Print WAAM testbed}
    \label{fig:testbed}
\end{figure}


Robotic WAAM processes present several challenges, with the foremost being geometry imprecision, which compromises near-net-shape capability, especially when working with low-melting-point metals such as aluminum alloys \cite{he2024opensourcesoftwarearchitecturemultirobot}.  
When operating commercial WAAM systems in the open-loop mode, defects can arise intermittently. Addressing these variations is essential to prevent error accumulation, which can compromise manufacturing quality. Some methods model the process using Gaussian Process Regression Models \cite{Xiong2019process,Dharmawan2020model} and apply reinforcement learning for optimization. Others incorporate additional control actions, such as arc control combined with milling \cite{ma2019optimization}, or use milling strategies to mitigate error accumulation \cite{LIM2022hybrid}. However, these approaches are less efficient and increase material usage because of the post-processing as compared to direct process improvement. Many of these methods rely on laser scanners to capture weld geometry profiles, while others utilize alternative data sources, such as acoustic signals \cite{surovi2021study,surovi2023acoustic}, for defect detection, enabling real-time correction and improved efficiency.

A robotic WAAM system typically involves a 6-dof welding robot combined with a 2-dof trunnion to enable alignment of metal deposition with gravity \cite{kim1998robot}.  An addition 6-dof robot mounted with sensors provides  comprehensive inspection coverage.
Motion of these robotic devices need to be coordinated to ensure task execution and avoidance of constraints such as joint limits and collision. 
Planning techniques to avoid robot singularities \cite{ahmad1989coordinated} and selection of welding paths \cite{bhatt2021optimizing} are commonly employed. The standard approach involves preplanned paths with consideration of robot kinematics \cite{Schmitz2021robot,Michel2019modular}. Although many methods incorporate redundancy for added flexibility, effective strategies are needed to resolve it \cite{Lizarralde2022online,hu2015task}. To address these challenges, we consider a three-robot setup, including a 6-dof welding robot, a positioner with 2-dof trunnion platform and a 6-dof sensing robot with a wrist-mounted laser line scanning measuring the printed height profile. We use the motion planning approach for multiple robot arm from our previous work \cite{he2023highspeed, he2024fast} for robot arms path planning and \cite{he2024opensourcesoftwarearchitecturemultirobot} gravity alignment. 

This paper presents a multi-robot framework for WAAM process monitoring and control. In the planning phase, the welding and inspection motion between the robots and the positioner are generated to meet the welding and inspection requirements. During the welding phase, we consider the robot torch path speed as the control input while holding the welding parameters, such as the wire feed rate at constant based on the materials used. The inspection robot measurements the layer height using a wrist-mounted laser scanner. The  scans from the laser scanner are registered using the calibrated robot kinematics, followed by  removal of outlier points to generate the height profile of the deposit. The torch speed for the next layer is then adjusted by the deviation from the desired profile. The adjustment is based on an identified model correlating the torch speed to the change in deposition height.  The control architecture, described in \cite{he2023plug}, coordinates the motion and data acquisition between all robots and sensors.  The closed loop WAAM framework has been applied in our testbed for printing a vertical wall, a complex turbine-blade-like geometry and a cylindrical geometry, 
\ifresubmissionred{\color{red}\fi 
using an aluminum alloy. 
\ifresubmissionred}\fi 
The geometric fidelity of these parts outperform the open-loop baselines in both layer-by-layer and continuous-scan modes. 
\ifresubmissionred{\color{red}\fi
We also apply our approach to a steel alloy, known for its higher melting point and greater thermal stability compared to aluminum alloys. The proposed techniques works well and shows improvements over the open loop printing.
\ifresubmissionred}\fi
We demonstrate the repeatability of the manufacturing process, so that once the corrected motion profile is learned, the scanning step may be removed, thereby enhancing manufacturing efficiency.
This paper is organized to present the scan-n-print framework in Section~\ref{sec:framework}, the deposition model in Section~\ref{sec:model}, and experimental results and evaluation in Section~\ref{sec:results}. 
The source code, datasets and video related to this study are publicly available.\footnote{\url{https://github.com/rpiRobotics/Convergent_Manufacturing_WAAM/tree/main/scan_n_print}} \footnote{\url{https://youtu.be/Ei2e2mBJZ5s}} 


\section{Scan-n-Print Framework}
\label{sec:framework}

\subsection{Overview}

\begin{figure}[ht]
    \centering
    \includegraphics[width=0.48\textwidth]{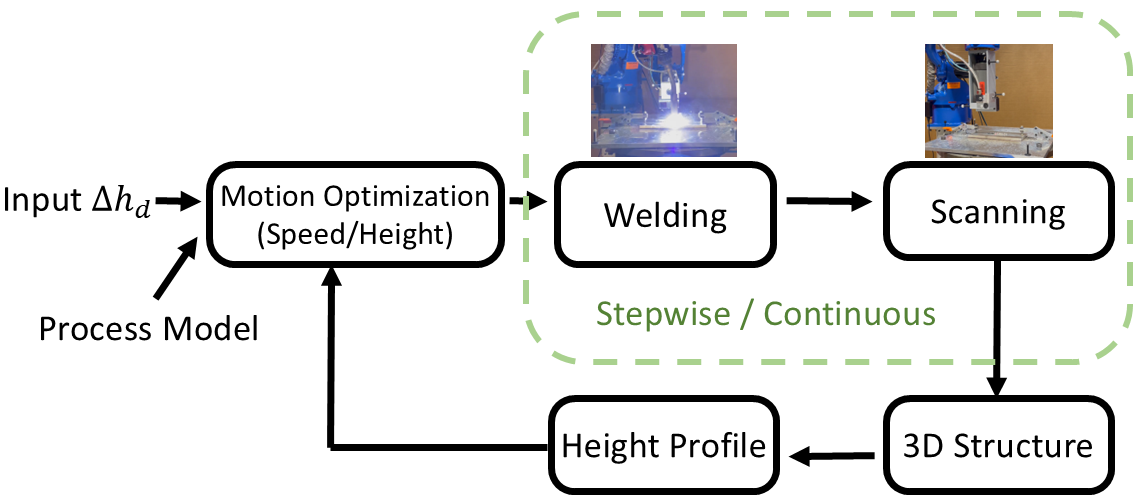}
    \caption{Scan-n-Print framework}
    \label{fig:weld_scan_framework}
\end{figure}

The proposed closed-loop WAAM framework is  illustrated in Figure~\ref{fig:weld_scan_framework}. 
After the deposition of each layer and scanning, the part geometry is reconstructed, with each scan registered to the positioner frame using the forward kinematics of the scanning robot and the positioner. Sampling the surface along the welding path gives the height profile of the top surface.  
For each layer, the slicing algorithm generates the desired layer height $\Delta h_d$.  The measured height profile of the current layer is compared with the desired profile and the difference, together with the torch speed to height model, generates the robot motion for the next layer. 

\subsection{Motion Generation and Execution}
\label{sec:motion_gen}

\begin{figure}[ht]
    \centering
    \includegraphics[width=0.47\textwidth]{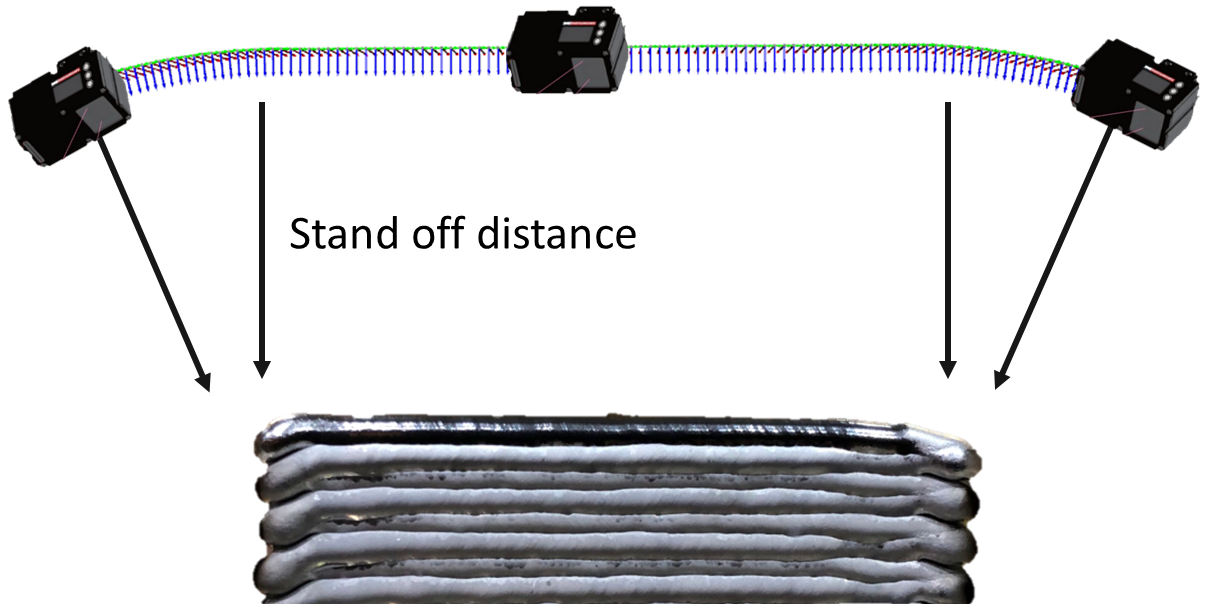}
    \caption{Scanning path plan}
    \label{fig:scanning_path}
\end{figure}

Motion planning is required for both the WAAM system (6-dof welding robot and 2-dof trunnion) and the sensing robot.  
As described in \cite{he2024opensourcesoftwarearchitecturemultirobot}, motion of the combined 8-dof system is determined by the torch position based on the slicing of the target geometry, the commanded path velocity, and the alignment requirement of the gravity direction with the previous layer.  There is one redundant degree of freedom for the rolling motion along the torch.  This redundancy resolved using resolved motion control subject to speed and joint limit constraints. 
%
%
%
The scanning motion of the sensing robot is designed to ensure complete coverage of the top layer. As illustrated in Figure~\ref{fig:scanning_path}, the scan path is  along the known welding trajectory while maintaining the appropriate laser-scan standoff distance. To fully cover critical areas at the start and end of the weld, side scanning motion is also used.  As for the welding motion, the scanning path is designed in the workpiece frame and then transformed to the positioner frame once the workpiece is placed on the positioner. The robot motion is designed to ensure 
that the scans plane is perpendicular to the deposited layer.  Since the gravity alignment is not needed, there is two 
redundant degrees of freedom (6-dof scanning robot and the 2-dof positioner). The same resolved motion control is used for redundancy resolution. 

\ifresubmissionred{\color{red}\fi
The overall system architecture to integrate multi-robot control and multi-sensor data acquisition
is described in 
\cite{he2024opensourcesoftwarearchitecturemultirobot}.
In the step-wise mode, the Cartesian motion of both the welding and scanning robots are segmented into uniform linear segments and encoded in the Motoman robot programming language, INFORM. The robot program together with the corresponding welding commands are sent to the Motoman DX200 robot controller. The DX200 controller coordinates the motion of the welding and scanning robots and trunnion table, and the Fronius weld controller. In the continuous mode, we bypass the Motoman controller and directly command the robot motion and weld controller. This is accomplished with two custom drivers using Robot Raconteur (RR) as the middleware \cite{rr2023wason}.  The robot driver is developed using the  Motoman MotoPlus package to stream robot motion setpoints and read robot joint angles at 8~ms to all three robots. The weld controller driver interfaces to the Fronius weld controller to change weld parameters and turn the welding arc on and off at 10 Hz.
\ifresubmissionred}\fi 

\subsection{Deposition Height Profile}
\label{sec:framework_height_profile}

\begin{figure}[ht]
    \centering
    \includegraphics[width=0.48\textwidth]{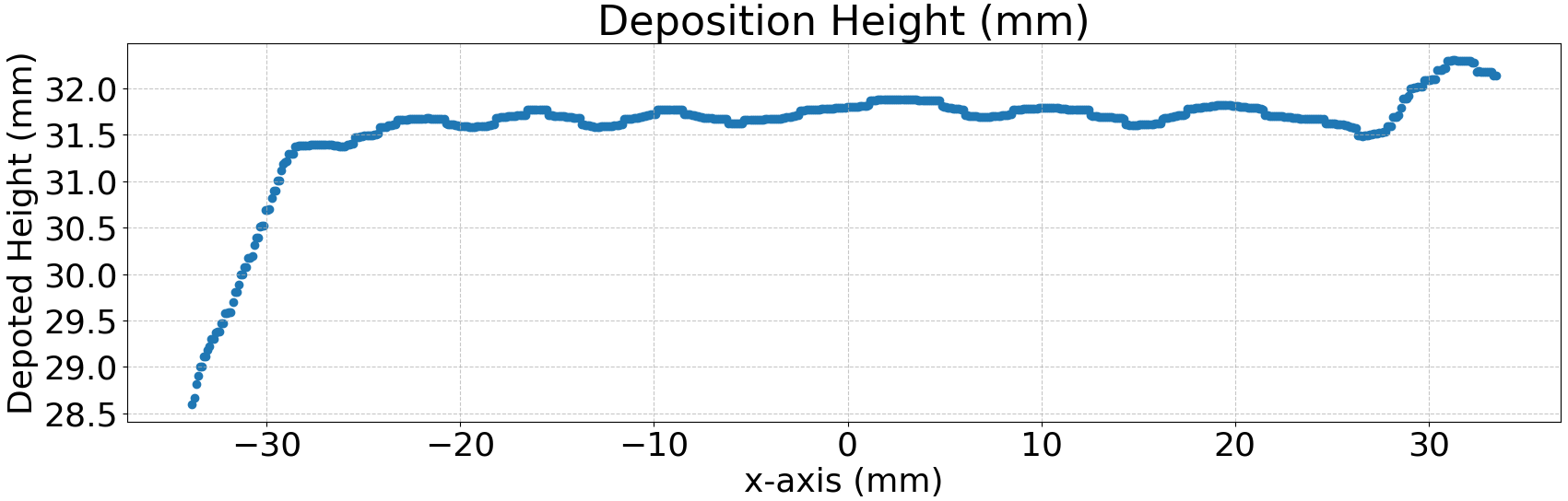}
    \caption{Deposition height of the weld piece}
    \label{fig:height_profile}
\end{figure}

During run time, each layer is scanned immediately after printing.
The laser scans 
form a 3D point cloud in the laser scanner frame.  To transform the points to the robot frame, we used the robot joint angles and the inverse kinematics which requires   
robot kinematic calibration \cite{xu2022handeye}. To mitigate noise, we apply statistical and cluster-based outlier removal, and a average smoother along the welding direction. The cleaned point cloud enables extraction of the weld top profile by sampling along welding travel direction. 
With a well-defined point cloud representation of the weld, 
we extract the layer deposition profile by slicing the point cloud along the printing direction. The height of each location is determined by the height of the points within the point cloud. Figure~\ref{fig:height_profile} displays the top layer height profile of the welded piece in the $x$-axis direction.

\subsection{Motion Optimization}

The motion optimization module aims to achieve uniform deposition height and correct welding defects.  It is an essential step to prevent error accumulation and ensure consistent surface quality. 
We adjust only the torch path speed while maintaining constant wire feed rate to regulate the amount of deposited materials at each time.  
Denote relationship between the torch speed and the weld deposition height at a given feed rate $\Delta h = f(v)$, where $v$ is the torch speed and $\Delta h$ the deposition height.  
Empirical characterization of this model is described in Section~\ref{sec:model}. In general, $f$ is a monotonically decreasing function as higher speed means less time for deposition and therefore smaller $\Delta h$.

\begin{figure}[ht]
    \centering
    \includegraphics[width=0.48\textwidth]{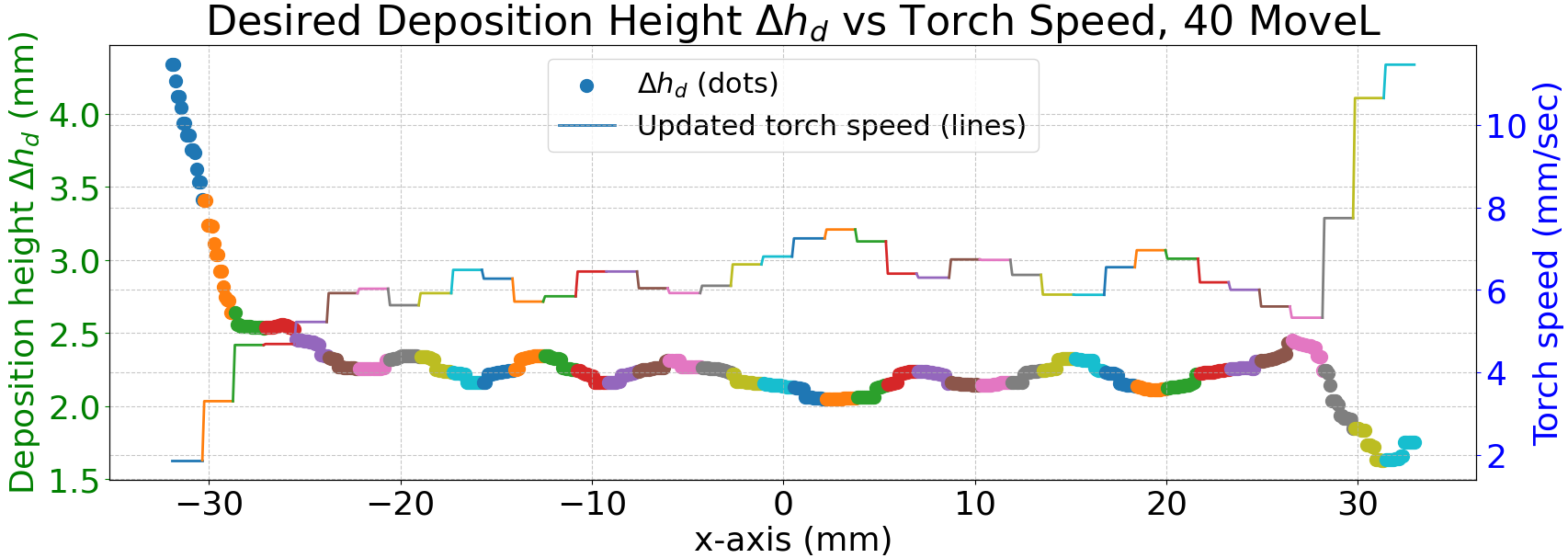}
    \caption{Speed update of 40 motion segments. Dots: difference from the current height to the target height $\Delta h_d$. Lines: the updated torch speed.}
    \label{fig:motion_update}
\end{figure}

Denote the average height of the current layer, $h^{(i)}$, calculated from the deposition profile. The target height for the next layer, $h^{(i+1)}_d$, is $h^{(i+1)}_d = h^{(i)} + \Delta h_d^{(i)}$, 
where $\Delta h_d^{(i)}$ is the desired deposition height.  
The robot motion for each layer is divided into linear torch motion segments. 
Let the average measured height for the $k$th segment be $\bar h^{(i,k)}$.
The desired deposition height for this segment is then calculated as $\Delta h^{(i,k)}_d = h^{(i+1)}_d - \bar{h}^{(i,k)}$.  Since the deposition model is monotonic, it is invertible. The inverse model then provides the required path speed:
\begin{equation}
    v^{(i,k)} = f^{-1}\left(\Delta h^{(i,k)}_d\right).
\end{equation}
Figure \ref{fig:motion_update} 
shows the measured heights and the corresponding commanding speeds for the motion segments along a typical layer. 
\ifresubmissionred{\color{red}\fi
%
We apply a moving average smoothing filter to the height profile. This is a low pass filter that remove high frequency components in the profile that could cause unwanted high acceleration.
The maximum and minimum speeds are also capped to prevent exaggerated movements. If a desired speed is not achieved because of the limited acceleration and speed, the remaining height errors are corrected in the subsequent layers.
\ifresubmissionred}\fi  
\ifresubmissionred{\color{red}\fi 
The optimized motion speed is commanded in the robot motion program if operated in the step-wise mode. In the continuous mode, we directly update the torch speed by adjusting the robot joint trajectory streamed to the robot controller in real time.
\ifresubmissionred}\fi

\subsection{Step-wise vs Continuous Scan-n-Print}
\label{sec:stepwise_continous}

The scan-n-print framework operates in step-wise or continuous modes. In the step-wise mode, robots alternate between welding and scanning, with the deposition profile processing and motion optimization occurring after the scanning motion. In the continuous mode, the scanning robot provides real-time feedback by looking ahead of the layer height, allowing on-the-fly processing and optimization. The continuous mode offers more immediate feedback and correction, and does not require turning the welding arc on and off (which compromises deposition uniformity) between layers. However, the implementation of the continuous mode is more complex due to the coordinated motion between all three robots
\ifresubmissionred{\color{red}\fi
The motion optimization method is the same in both modes, but the update frequency is different. The step-wise mode optimizes the motion of the entire layer, while the continuous mode optimizes the motion of the current welding point during run-time using the look-ahead scanner. In our setup, the continuous mode closes the feedback loop at 100~Hz, limited by the scanning rate. Both modes produce high quality final parts.  We observe that the continuous mode has slightly better performance in terms of the height standard deviation.  This is because the step-wise mode requires turning the welding arc on and off at the edges. 
%
\ifresubmissionred}\fi 

\section{Welding Model}
\label{sec:model}


To derive the model linking welding motion speed to deposition height using the collected layer data, we reference the deposition equation in \cite{Handa1997THERE}:  \begin{equation} 
    S_w v_{MR} = S_B v 
\end{equation} 
where $S_w$ is the wire cross-section, $v_{MR}$ is the wire melting rate, $S_B$ is the weld bead cross-section, and $v$ is the welding torch speed. Note that $v_{MR}$ is proportional to the wire feed rate. 
Assuming that the cross-section of the weld bead width is proportional to the height with proportionality constant $c$, we have approximately
\begin{equation} 
    \frac{S_w v_{MR}}{c\Delta h^2} = v .
\end{equation} 
It follows that 
\def\half{\frac12}
\begin{equation} 
\ln(\Delta h) = -\half \ln(v) + \half \ln\left(\frac{S_w v_{MR}}{c}\right) 
\end{equation} 
To account for inaccuracy in the model, including the complex weld shape, we consider a more general expression:
\begin{equation} 
    \ln(\Delta h) = a \ln(v) + b 
\label{eq:model_equation}
\end{equation} 
where $a$ and $b$ are constants that depend on the wire feed rate, to be determined based on collected data.

\begin{figure}[ht]
    \centering
    \includegraphics[width=0.45\textwidth]{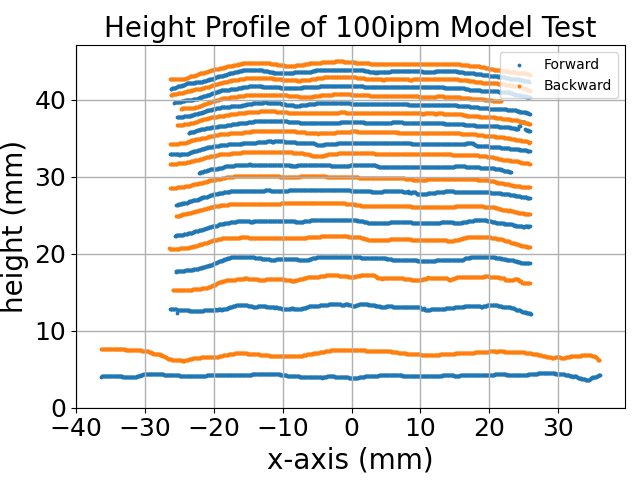}
    \caption{100 ipm layers height. The speed of the torch varies from 2 mm/sec at the bottom layers to 20 mm/sec at the top layers to collect deposition data at each traveling speed. }
    \label{fig:100ipm_height}
\end{figure}

\begin{figure}
    \centering
     \includegraphics[width=0.45\textwidth]{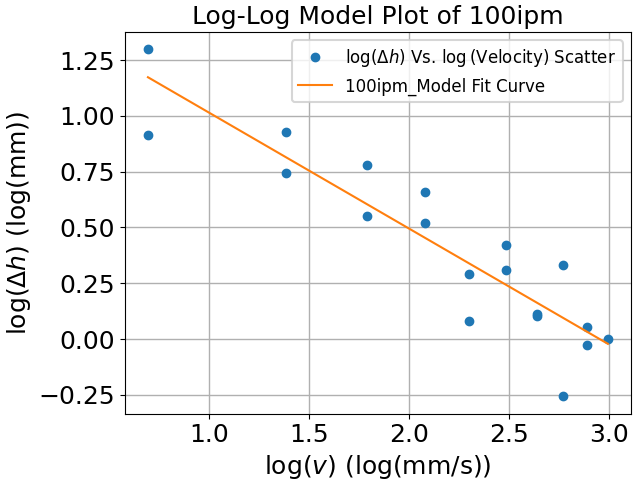}
     \caption{Welding motion speed to deposition model for 100 ipm wire feed rate. Table.~\ref{tab:model_param} shows the model parameters of all tested wire feed rate with the data fitting RMSE.
     }
     \label{fig:100ipm_model}
\end{figure}

To identify the model, we collect data as follows:
\begin{itemize}
[topsep=4pt,itemsep=0pt,partopsep=4pt, parsep=0pt,leftmargin=*]
    \item Weld two base layers for a consistent foundation for each run.
    \item Reduce the torch speed  by 2mm/sec every two layers, from 20mm/sec  to 2mm/sec, while maintaining a constant feed rate.
    \item Measure the average height of each welded layer.
\end{itemize}
Figure \ref{fig:100ipm_height} shows the layer heights of a weld sample collected under fixed wire feed rate at 100~ipm (inch per minute).  As expected, the graph shows that an increase in torch speed results in a decrease in deposition height. Figure~\ref{fig:100ipm_model} presents the identified model parameters, with least square fit of a line to the collected data.
By repeating this procedure with varying feed rates, model functions for different wire feed rates can be developed.

The identified coefficients are provided in Table~\ref{tab:model_param}. Note that the coefficients of the first-order terms are consistently very close to $-0.5$ which aligns with the equation derivation. We also showed the RMSE of the data after fitting the model. \ifresubmissionred{\color{red}\fi
During the scan-n-print operation, the wire feed rate is fixed at a chosen value and the torch speed is adjusted based on the scanning result.
\ifresubmissionred}\fi 
\begin{table}[ht]
\begin{center}
\begin{tabular}{ c c c c c c c } 
\toprule
feed rates& 100 & 110 & 130 & 150 & 170 & 240  \\
\midrule
a (slope) & -0.62 & -0.43 & -0.43 & -0.44 & -0.45  & -0.46 \\
b & 1.85 & 1.23 & 1.630 & 1.37 & 1.40 & 1.15 \\
RMSE & 0.27 & 0.12 & 0.13 & 0.18 & 0.21 & 0.10 \\
\bottomrule
\end{tabular}
\caption{Identified model parameters and regression error (in mm) under various constant wire feed rates (in inch-per-minute)}
\label{tab:model_param}
\end{center}
\end{table}

\section{Experiments and Evaluation}
\label{sec:results}

\subsection{Hardware Testbed}

The hardware testbed is shown in Figure \ref{fig:testbed}. It comprises a MA2010 welding robot with a D500B positioner, and a MA1440 inspection scanning robot, all from Yaskawa Motoman.  Fronius 500i provides the welding power source for the cold metal transfer process.  The Fronius weld controller is integrated with the robot controller DX200.
The weld material is ER~4043, an aluminum alloy with lower melting point 
compared to materials like steel alloy or stainless steel. Its high thermal conductivity often leads to varying weld conditions and a higher occurrence of defects. \ifresubmissionred{\color{red}\fi
In Section~\ref{sec:steel_material}, we apply scan-n-print to ER~70S-6, a steel alloy with higher melting point and lower thermal conductivity than the aluminium alloy.
\ifresubmissionred}\fi 
For laser scanning, we use an MTI Pro-Track~G Series laser scanner installed on the inspection robot. The scanner must maintain a standoff distance between 65mm to 125mm with a field of view of approximately 40mm to 60mm. The scans are collected at a frame rate of about 100~Hz.
%
\begin{figure}[ht]
    \centering
    \includegraphics[width=0.48\textwidth]{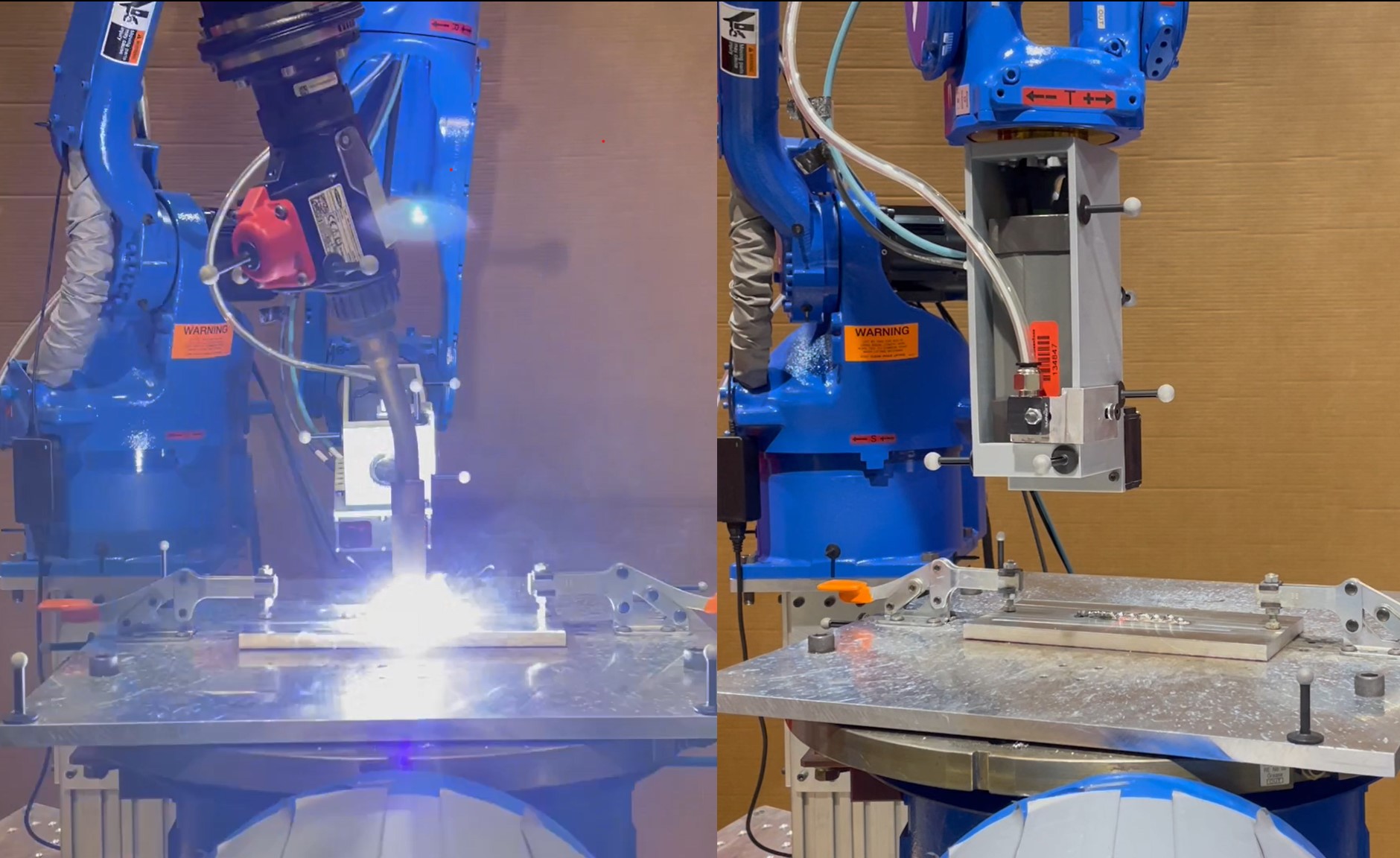}
    \caption{Robotics WAAM and scanning execution}
    \label{fig:weld_scan}
\end{figure}

\subsection{Performance Comparison: Wall Structure}
\label{sec:exp_wall}

For the baseline, we select the best combination of wire feed rate and torch speed: 100~ipm and 5mm/sec.
With the identified model in Section~\ref{sec:model}, the input height is approximately $\Delta h_d \approx 2.34$mm when the torch speed is set to 5mm/sec.  For comparison between the scan-n-print and open loop, we print a wall structure with width $65$mm and thickness $4$mm, and the top layer exceeding 50mm in height.
\begin{figure}[htbp]
     \centering
     \begin{subfigure}[b]{0.21\textwidth}
         \centering
         \includegraphics[width=\textwidth]{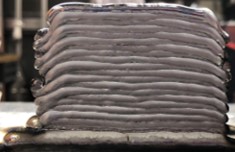}
         \caption{Open-loop Baseline}
     \end{subfigure}
     \begin{subfigure}[b]{0.23\textwidth}
         \centering
         \includegraphics[width=\textwidth]{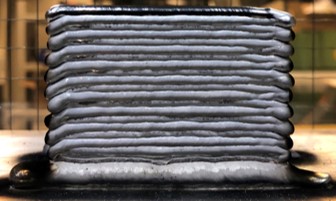}
         \caption{Scan-n-Print framework}
     \end{subfigure}
    \caption{Open-loop baseline vs. Scan-n-Print for a Wall Structure, \ifresubmissionred{\color{red}\fi
    ER~4043 Aluminum Alloy.
\ifresubmissionred}\fi}
    \label{fig:weld_piece}
\end{figure}

\begin{figure}[htbp]
     \centering
     \begin{subfigure}[b]{0.235\textwidth}
         \centering
         \includegraphics[width=\textwidth]{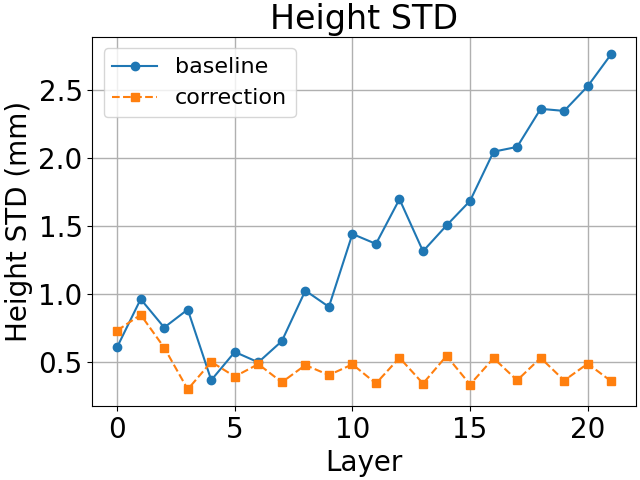}
         \caption{Full piece}
         \label{fig:height_std_full}
     \end{subfigure}
     \begin{subfigure}[b]{0.235\textwidth}
         \centering
         \includegraphics[width=\textwidth]{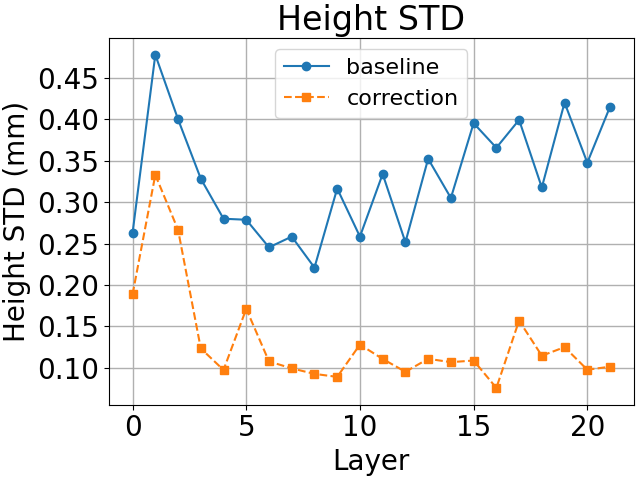}
         \caption{Edge excluded}
         \label{fig:height_std_cut}
     \end{subfigure}
    \caption{Height STD of each layer of the WAAM wall. Open-loop baseline vs closed-loop correction, \ifresubmissionred{\color{red}\fi
    ER~4043 Aluminum Alloy.
\ifresubmissionred}\fi}
    \label{fig:height_std}
\end{figure}

Figure \ref{fig:weld_piece} shows the printed pieces produced for the baseline and scan-n-print.  Even through visual observation alone, it is evident that the quality of the printed piece has significantly better layer uniformity with the closed-loop approach. Note that defects tend to occur predominantly at the edges of the weld piece due to the arc on/off effects.
Figure \ref{fig:height_std_full} shows the standard deviation of the layer height for both the open-loop baseline and closed-loop correction tests. In the open loop case, the height variation in each layer
accumulates with the number of layers.  
To evaluate the print performance without the edge effects, 
we exclude the edge data. \ifresubmissionred{\color{red}\fi
We define the edges as within $7.5$mm from the welding arc on and off points. We can identify the regions on the point cloud using the collected robot joints data and the associated scans.
\ifresubmissionred}\fi 
Figure~\ref{fig:height_std_cut} shows that 
the closed-loop approach still outperforms the open-loop approach. 
Table \ref{table:closed_loop_wall_performance} shows the statistics and improvements of the height uniformity over all layers. For the wall geometry, the standard deviation of height improved 66\%.  Even without the edges where defects dominate, the performance improved by 60\%.
\begin{table}[ht!]
\centering
\begin{tabular}{c c c}
\toprule
Mean Height STD & With edge & Without edge \\
\midrule
Baseline (mm) & 1.38 & 0.33 \\
Correction (mm) & 0.47 & 0.13 \\
Improvement (\%) & 66 & 60 \\
\bottomrule
\end{tabular}
\caption{Mean height standard deviations across all layers, with and without the edges, of the WAAM wall, for open-loop baseline and scan-n-print correction cases, \ifresubmissionred{\color{red}\fi
    ER~4043 Aluminum Alloy.
\ifresubmissionred}\fi }
\label{table:closed_loop_wall_performance}
\end{table}

\begin{figure}[htbp]
     \centering
     \begin{subfigure}[b]{0.20\textwidth}
         \centering
         \includegraphics[width=\textwidth]{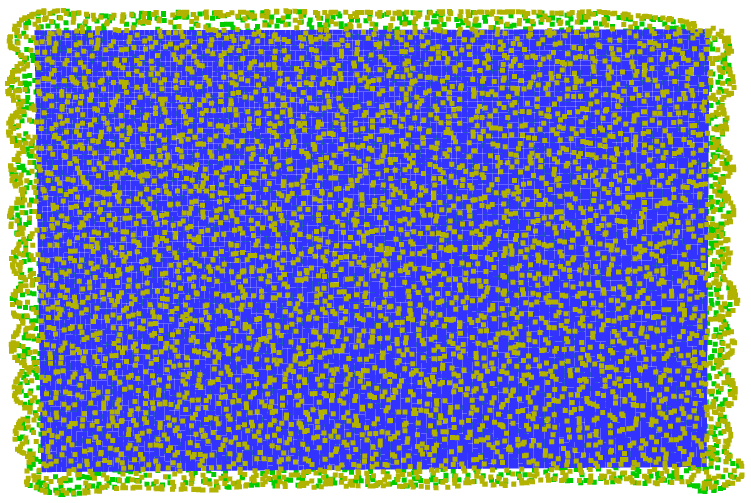}
         \caption{Wall 3D profile}
         \label{fig:wall_pcd}
     \end{subfigure}
     \begin{subfigure}[b]{0.26\textwidth}
         \centering
         \includegraphics[width=\textwidth]{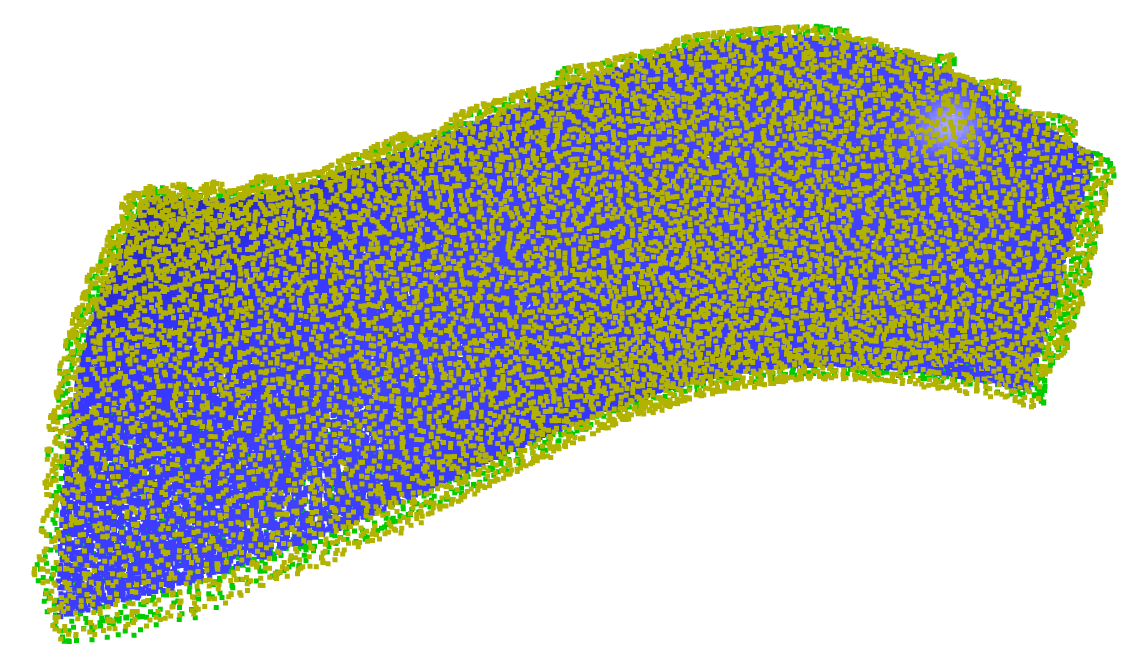}
         \caption{Blade 3D profile}
         \label{fig:blade_pcd}
     \end{subfigure}
    \caption{The 3D profile of the printed wall and blade (point cloud data from the Artec scanner) vs. the desired structure CAD model (blue). }
    \label{fig:3d_pcd}
\end{figure}

\begin{figure}[htbp]
     \centering
     \begin{subfigure}[b]{0.215\textwidth}
         \centering
         \includegraphics[width=\textwidth]{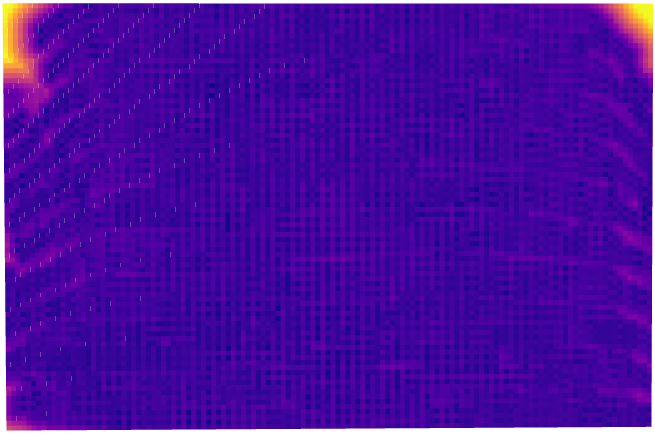}
         \caption{Baseline}
         \label{fig:3d_error_baseline}
     \end{subfigure}
     \begin{subfigure}[b]{0.255\textwidth}
         \centering
         \includegraphics[width=\textwidth]{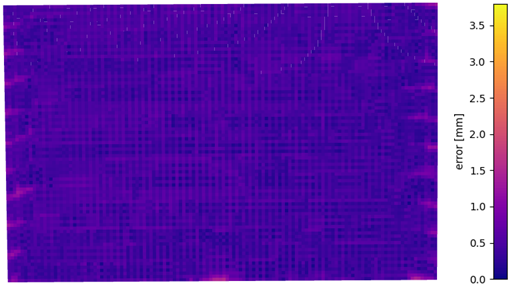}
         \caption{Correction}
         \label{fig:3d_error_correction}
     \end{subfigure}
    \caption{Error heat-map between the desired wall structure CAD model and the WAAM wall.}
    \label{fig:3d_error}
\end{figure}

We further illustrate the accuracy comparison of the printed WAAM wall structure versus the desired CAD model as shown in Figure~\ref{fig:wall_pcd}. The WAAM part is scanned using the Artec Spider scanner to generate a point cloud representation.  It is then aligned with the CAD model using Iterative Closest Point (ICP) registration \cite{besl1992icp}. The accuracy is characterized by the Euclidean distance error between the printed WAAM part and the desired structure. Figure~\ref{fig:3d_error} shows the error heat map on the wall CAD model. The heat map reinforces the observation that the proposed framework is far more accurate than the baseline. The error is mostly due to the defect on the edges. Table \ref{table:3D_profile_compare} shows the error statistics. The maximum error improves by 70\% and the average error improves by 13\%.

\begin{table}[ht!]
\centering
\begin{tabular}{c c c}
\toprule
Wall & Average Error & Max error \\
\midrule
Baseline (mm) & 0.39 & 5.38 \\
Correction (mm) & 0.34 & 1.52 \\
Improvement (\%) & 13 & 72 \\
\bottomrule
\end{tabular}
\caption{
Average and maximum distance between the WAAM wall and the CAD model}
\label{table:3D_profile_compare}
\end{table}


\subsection{Performance Comparison: Blade-like Structure}

For a more complex geometry, 
we apply the proposed framework to a generic fan blade geometry. The experiment was conducted following the same setup as the wall structure, with the wire feed rate at 100 ipm and desired deposition height of $\Delta h_d = 2.34$mm. 
\begin{figure}[htbp]
     \centering
     \begin{subfigure}[b]{0.22\textwidth}
         \centering
         \includegraphics[width=\textwidth]{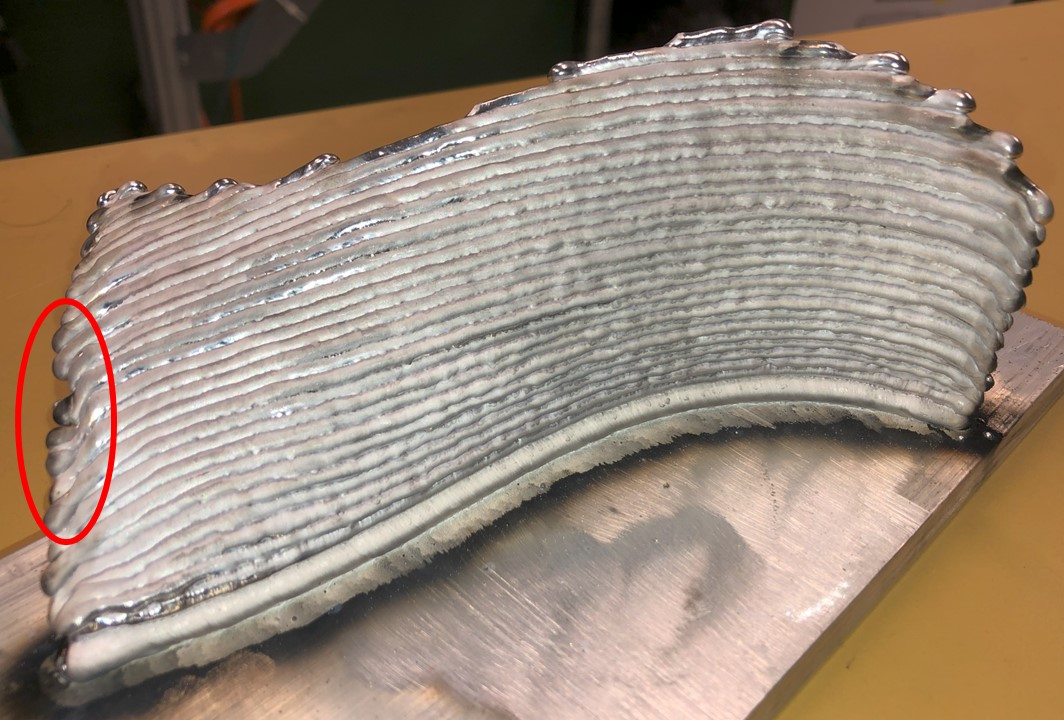}
         \caption{Open-loop WAAM blade}
         \label{fig:blade_baseline}
     \end{subfigure}
     \begin{subfigure}[b]{0.25\textwidth}
         \centering
         \includegraphics[width=\textwidth]{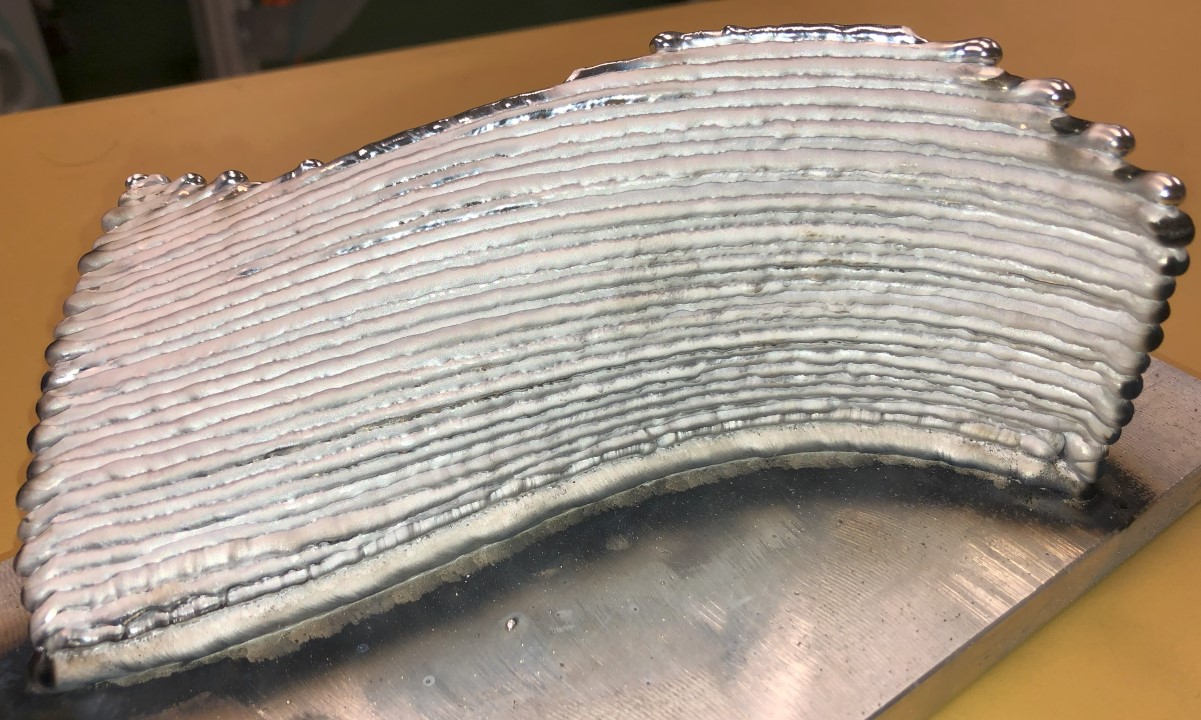}
         \caption{Scan-n-Print WAAM blade}
         \label{fig:blade_correction}
     \end{subfigure}
    \caption{A generic fan blade geometry using the proposed scan-n-print framework. The red circle shows the defect when using the open-loop setting.}
    \label{fig:blade_scanprint}
\end{figure}

\begin{figure}[htbp]
     \centering
     \begin{subfigure}[b]{0.235\textwidth}
         \centering
         \includegraphics[width=\textwidth]{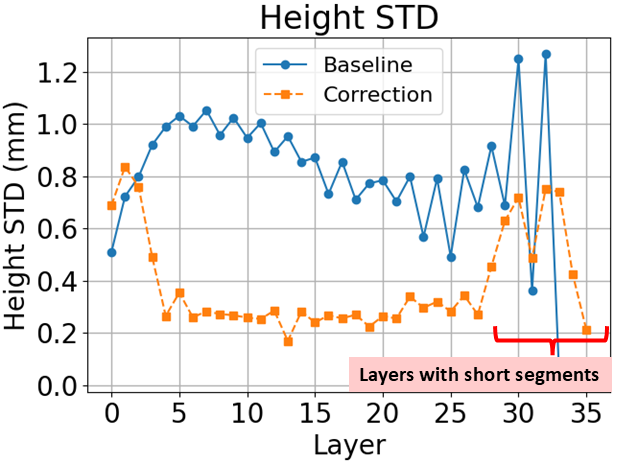}
         \caption{Height STD}
         \label{fig:blade_std}
     \end{subfigure}
     \begin{subfigure}[b]{0.235\textwidth}
         \centering
         \includegraphics[width=\textwidth]{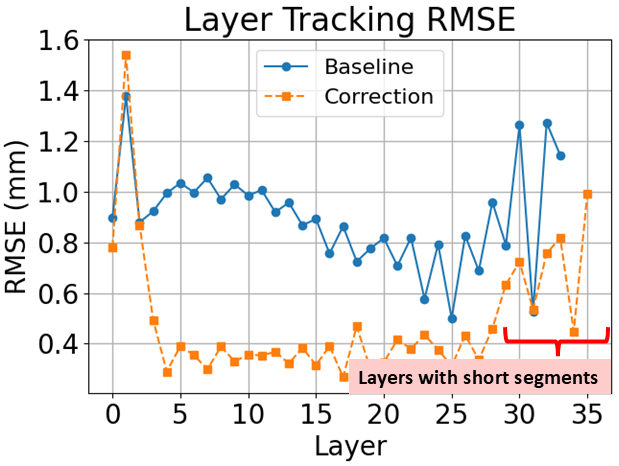}
         \caption{Height RMSE}
         \label{fig:blade_rmse}
     \end{subfigure}
    \caption{Height STD and RMSE of each layer of the WAAM blade}
    \label{fig:blade_height}
\end{figure}

Figure \ref{fig:blade_scanprint} shows the blade both in open loop and scan-n-print modes. 
A visual comparison reveals a clear defect at the edge (indicated by the red circle) in the open loop case, where the welding torch fails to deposit sufficient material at the edge of the piece. The closed-loop print shows a far smoother edge. 
Figure~\ref{fig:blade_std} shows the standard deviation and root mean square error (RMSE) of the deposition heights at each layer.  The larger error at the top layers is due to the short segments where the smaller number of measurements tend to accentuate these metrics.
Table \ref{table:closed_loop_blade_performance} shows the statistics and the improvements. The height STD improved by 53 \% and tracking rmse improved by 45 \%. 

\begin{table}[ht!]
\centering
\begin{tabular}{c c c}
\toprule
Blade & Mean Height STD  & Mean Tracking RMSE \\
\midrule
Baseline (mm) & 0.85  & 0.90 \\
Correction (mm) & 0.24  & 0.49 \\
Improvement (\%) & 53  & 45 \\
\bottomrule
\end{tabular}
\caption{Mean height STD and tracking RMSE across the printed layers of the WAAM blade.}
\label{table:closed_loop_blade_performance}
\end{table}

The comparison of the printed blade  and the blade CAD model is shown in Figure~\ref{fig:blade_pcd}.
Table~\ref{table:3D_profile_compare_blade} summarizes the error statistics. The maximum error improves by 5\% and the average error improves by 13\%.

\begin{table}[ht!]
\centering
\begin{tabular}{c c c}
\toprule
Blade & Average Error & Max error \\
\midrule
Baseline (mm) & 0.75 & 2.23 \\
Correction (mm) & 0.65 & 2.11 \\
Improvement (\%) & 13 & 5 \\
\bottomrule
\end{tabular}
\caption{Average and maximum distance between the WAAM blade and the CAD model}
\label{table:3D_profile_compare_blade}
\end{table}

\ifresubmissionred{\color{red}\fi
\subsection{Performance Comparison: Steel Alloy}
\label{sec:steel_material}

We follow the same procedure in Section~\ref{sec:exp_wall} but using the steel alloy ER~70S-6. We select the best combination of wire feed rate and torch speed for the steel alloy: 200 ipm and 7 mm/sec. We apply the same model identification procedure in Section~\ref{sec:model}. The input height is approximately $\Delta h_d \approx 1.35$mm when the torch speed is set to 7mm/sec. The wall structure has the same dimension with width 65~mm and thickness 4~mm.

\begin{figure}[htbp]
     \centering
     \begin{subfigure}[b]{0.21\textwidth}
         \centering
         \includegraphics[width=\textwidth]{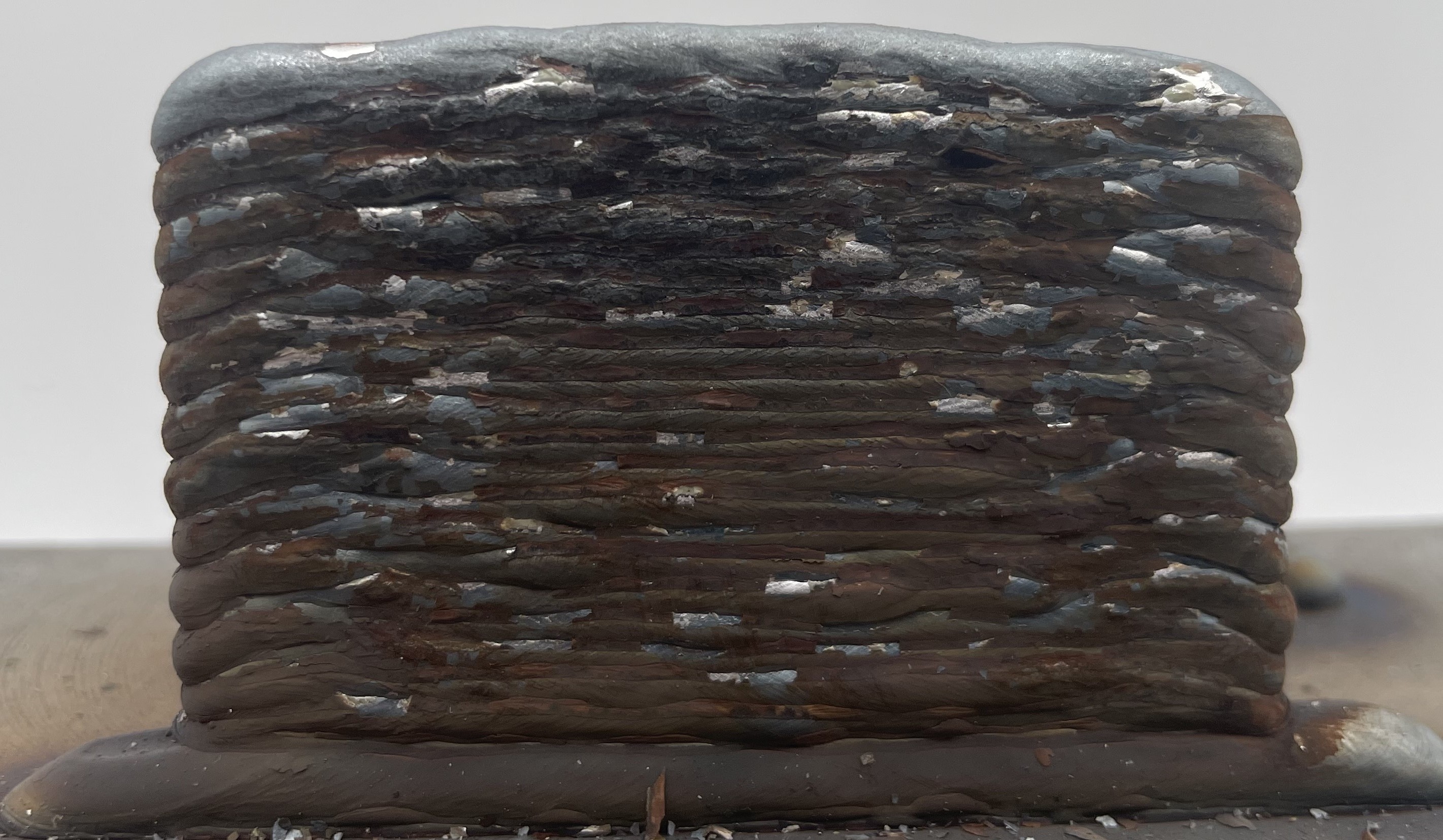}
         \caption{Open-loop Baseline}
     \end{subfigure}
     \begin{subfigure}[b]{0.23\textwidth}
         \centering
         \includegraphics[width=\textwidth]{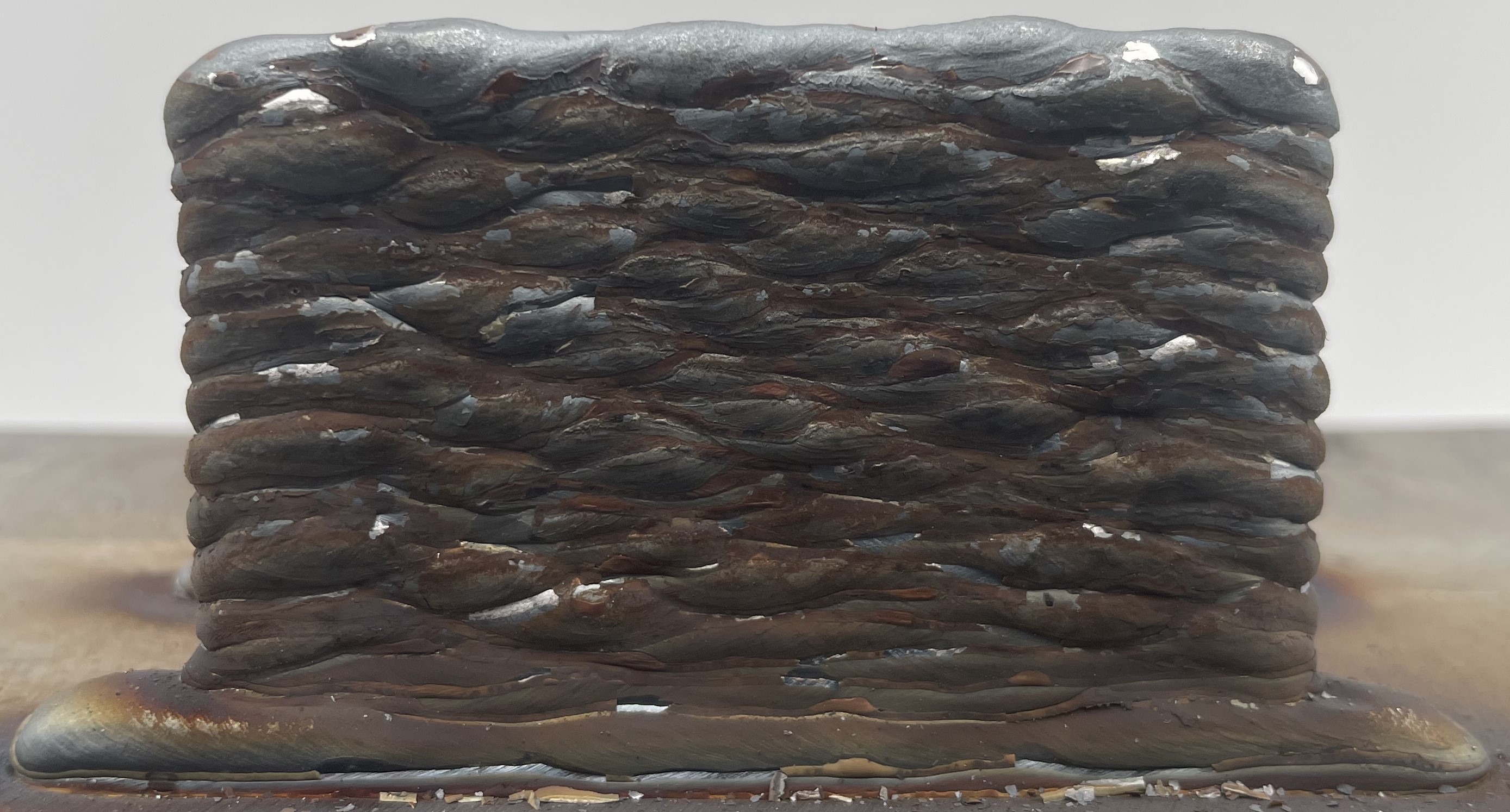}
         \caption{Scan-n-Print framework}
     \end{subfigure}
    \caption{\ifresubmissionred{\color{red}\fi
    Open-loop baseline vs. Scan-n-Print for a Wall Structure, ER~70S-6 Steel Alloy
\ifresubmissionred}\fi}
    \label{fig:weld_piece_steel}
\end{figure}

\begin{figure}[htbp]
     \centering
     \begin{subfigure}[b]{0.235\textwidth}
         \centering
         \includegraphics[width=\textwidth]{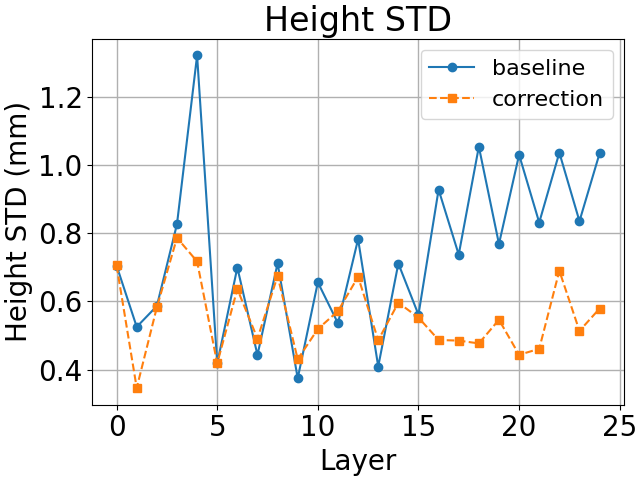}
         \caption{Full piece}
         \label{fig:height_std_full_steel}
     \end{subfigure}
     \begin{subfigure}[b]{0.235\textwidth}
         \centering
         \includegraphics[width=\textwidth]{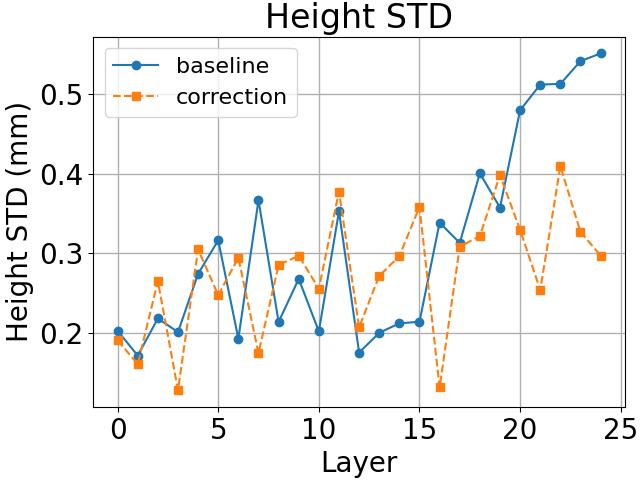}
         \caption{Edge excluded}
         \label{fig:height_std_cut_steel}
     \end{subfigure}
    \caption{\ifresubmissionred{\color{red}\fi 
    Height STD of each layer. Open-loop baseline vs closed-loop correction,  ER 70S-6 steel alloy.
    \ifresubmissionred}\fi}
    \label{fig:height_std_steel}
\end{figure}

Figure~\ref{fig:weld_piece_steel} shows the printed steel pieces for the baseline and scan-n-print. By visual observation, the open-loop welding with steel alloy does not create as much edge defects as with the aluminum alloy. This is because the steel alloy has higher melting point and lower thermal conductivity and is more stable than aluminum alloy. However, the top layer is still slightly curved compare to the close-loop approach. Figure~\ref{fig:height_std_full_steel} shows the standard deviation of the layer height for both the open-loop baseline and closed-loop correction tests. The open-loop case shows the error accumulation, but not as much as in the aluminum alloy case. The proposed scan-n-print still shows improvement. We exclude the edge data and evaluate the performance without edge effects. Figure~\ref{fig:height_std_cut_steel} shows that the uniformity between the open-loop and the closed-loop approach is comparable. However, we see a trend of error accumulation in the open-loop case at the end of the weld and the height STD exceeds the closed-loop case significantly. Table~\ref{table:closed_loop_wall_performance_steel} shows the statistics and improvements of the steel alloy case. As expected, the steel alloy is stable for WAAM so the mean height does not vary as much as the aluminum alloy in the open-loop baseline case (1.47 vs 0.74) but the proposed scan-n-print can further improve the weld by 26\%. Without considering the edges, the performance improves by 10\%.

\begin{table}[ht!]
\centering
\begin{tabular}{c c c}
\toprule
Mean Height STD & With edge & Without edge \\
\midrule
Baseline (mm) & 0.74 & 0.31 \\
Correction (mm) & 0.55 & 0.28 \\
Improvement & 26\% & 10\% \\
\bottomrule
\end{tabular}
\caption{
\ifresubmissionred{\color{red}\fi
    Mean height standard deviations across all layers, with and without the edges, of the WAAM wall, for open-loop baseline and scan-n-print correction cases, ER~70S-6 Steel Alloy. 
\ifresubmissionred}\fi}
\label{table:closed_loop_wall_performance_steel}
\end{table}

\ifresubmissionred}\fi 

\subsection{Continuous Scan-n-Print}

\begin{figure}[htbp]
     \centering
     \begin{subfigure}[b]{0.13\textwidth}
         \centering
         \includegraphics[width=\textwidth]{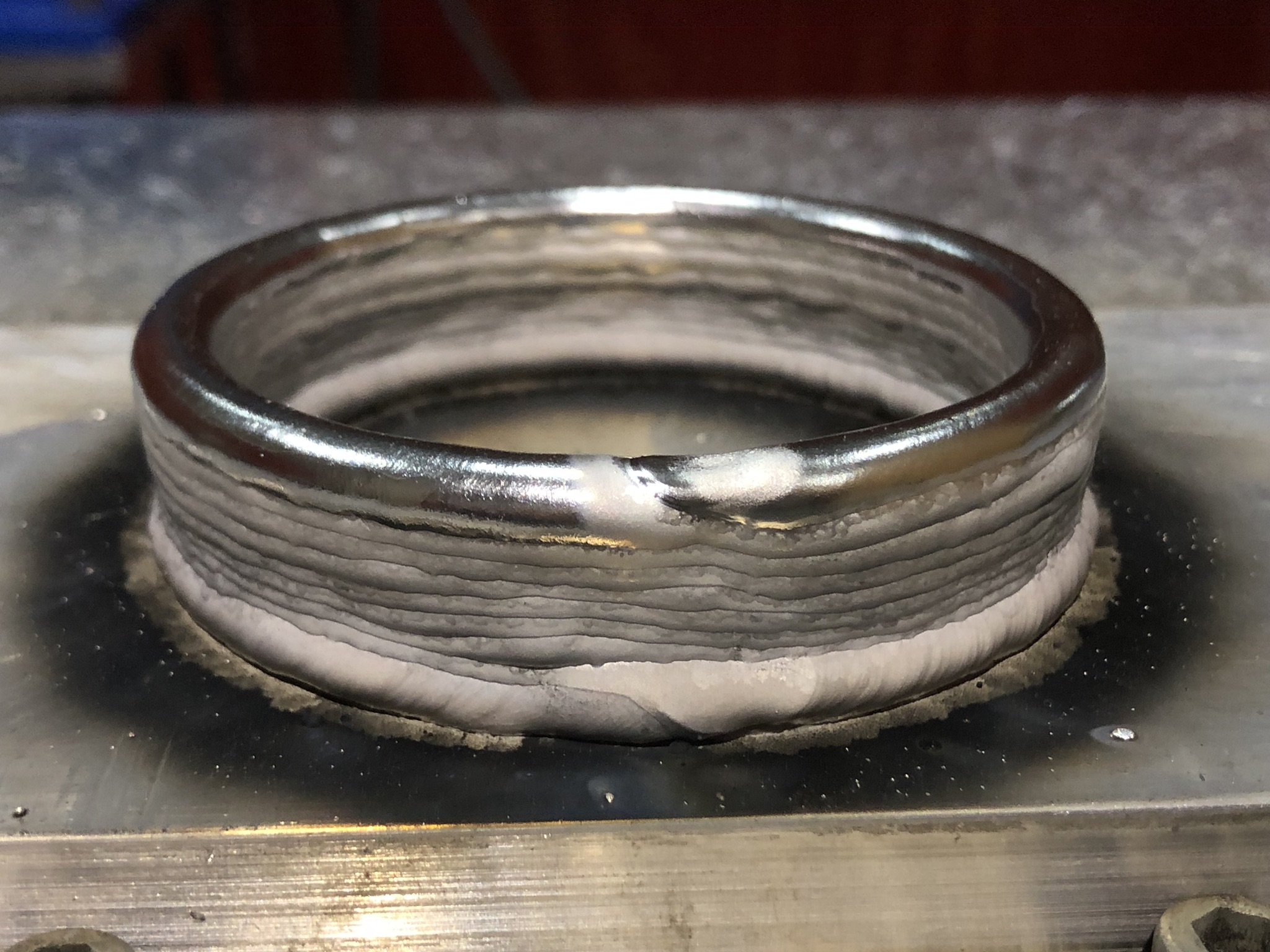}
         \caption{Cylinder}
     \end{subfigure}
     \begin{subfigure}[b]{0.16\textwidth}
         \centering
         \includegraphics[width=\textwidth]{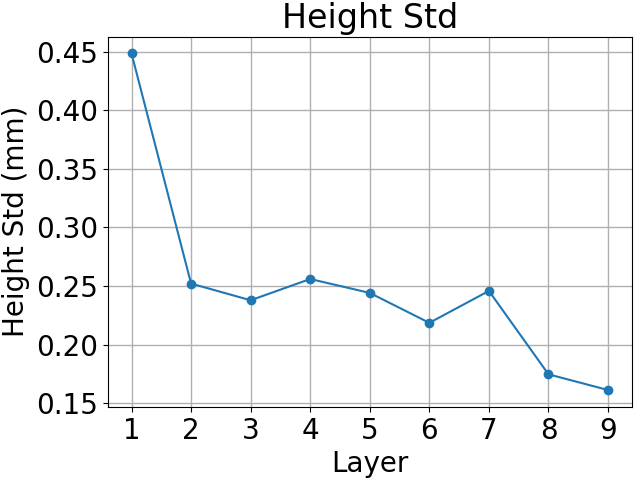}
         \caption{Height STD}
     \end{subfigure}
     \begin{subfigure}[b]{0.16\textwidth}
         \centering
         \includegraphics[width=\textwidth]{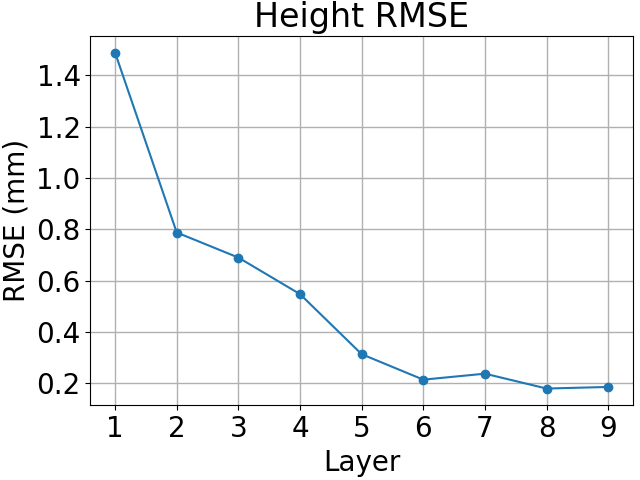}
         \caption{Height RMSE}
     \end{subfigure}
    \caption{Continuous Scan-n-Print of a cylinder. }
    \label{fig:cont_weld}
\end{figure}

We also demonstrated the scan-and-print framework in a continuous mode setup, with the same control algorithm, targeting a cylindrical geometry with a radius of $r=35$ mm. In this setup, the scanning robot is positioned half the perimeter ahead of the welding torch. The leading scanner provides look-ahead feedback by capturing the coming geometry surface and storing it in the memory. The torch speed is then adjusted while reaching the corresponding position.
The feed rate was set at 160 ipm with a desired deposition height of $\Delta h_d = 1.80$ mm. Figure \ref{fig:cont_weld} shows the results of the continuous scanning and printing process. The significant reduction in standard deviation demonstrates the effectiveness of the proposed framework in smoothing the height variation in each layer. 

\subsection{Repeatability Test}

\begin{figure}[htbp]
     \centering
     \begin{subfigure}[b]{0.15\textwidth}
         \centering
         \includegraphics[width=\textwidth]{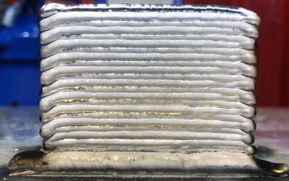}
         \caption{Recorded}
     \end{subfigure}
     \begin{subfigure}[b]{0.15\textwidth}
         \centering
         \includegraphics[width=\textwidth]{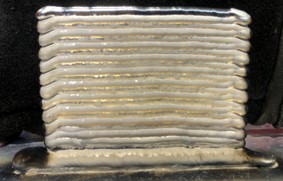}
         \caption{Repeat 1}
     \end{subfigure}
     \begin{subfigure}[b]{0.15\textwidth}
         \centering
         \includegraphics[width=\textwidth]{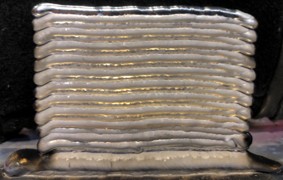}
         \caption{Repeat 2}
     \end{subfigure}
    \caption{Repeatability test.}
    \label{fig:repeat_test}
\end{figure}

\begin{figure}[htbp]
     \centering
     \begin{subfigure}[b]{0.235\textwidth}
         \centering
         \includegraphics[width=\textwidth]{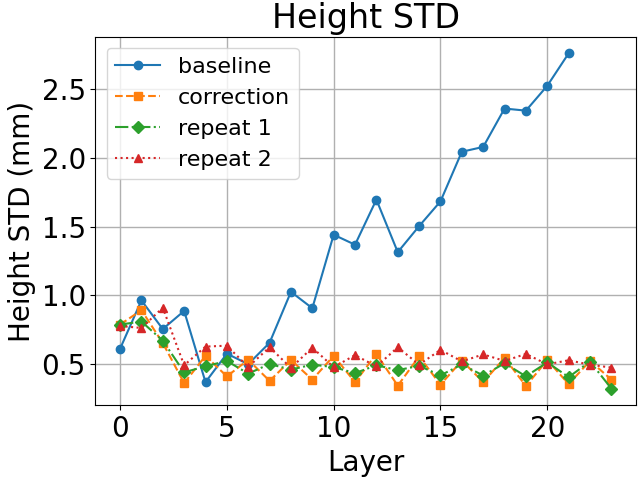}
         \caption{Full Piece}
         \label{fig:height_std_repeat_full}
     \end{subfigure}
     \begin{subfigure}[b]{0.235\textwidth}
         \centering
         \includegraphics[width=\textwidth]{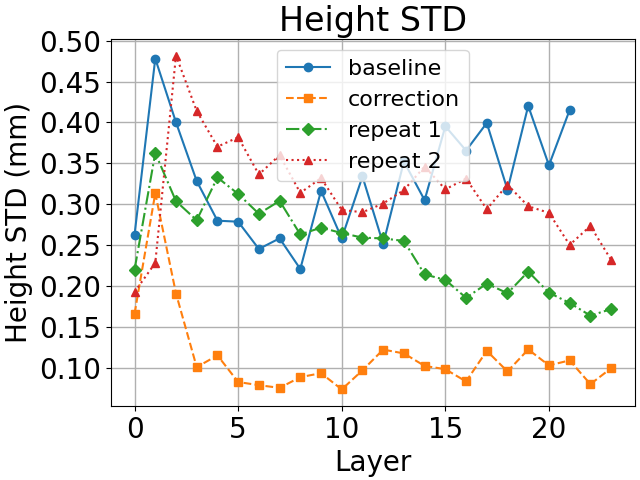}
         \caption{Edge Excluded}
         \label{fig:height_std_repeat_cut}
     \end{subfigure}
    \caption{Repeatability test: height STD of each layer. Open-loop baseline vs closed-loop approach vs repeats}
    \label{fig:height_std_repeat}
\end{figure}

For printing of multiple parts, we may use scan-n-print to determine the torch speed in each layer and then apply the recorded torch speed profiles in the more efficient open loop printing of subsequent parts.  
%
%
Figure \ref{fig:repeat_test} shows  the comparison between the part produced by the scan-n-print and parts produced by replaying the recorded velocity commands. Qualitatively, the replayed printing match closely the  original closed loop print, with the notable absence of the edge defects observed in the baseline part.  The top layer also appears to be relatively uniform. The standard deviation of layer heights, shown in Figure \ref{fig:height_std_repeat_full}, corroborates these observations. 
When the edge regions are excluded from the analysis, shown in Figure \ref{fig:height_std_repeat_cut}, 
the replayed prints perform similarly as the open loop, indicating that the correction is mostly for the edge of the wall.


\section{Conclusion} 

In this paper, we introduced a scan-n-print framework to improve the performance of the  WAAM processes. The height variation is measured using a laser scanner and the torch speed is adjusted to compensate for the height variation.  Layer height regulation is accomplished by inverting an identified model relating the torch speed to the deposition height.  
The effectiveness of the proposed approach is demonstrated on several printed geometries, vertical wall, a mock fan blade, and a cylinder.  
The demonstration is conducted on our WAAM testbed with a welder robot and positioner, and a separate monitoring robot with a mounted laser scanner.
We are currently working on regulating both the geometry and temperature by controlling both wire feed rate and torch speed and using an IR camera for temperature feedback in addition to the laser scanner.


\section*{ACKNOWLEDGMENT}
The authors would also like to thank Chris Anderson and Roger Chistian at Yaskawa Motoman, Jeff Schoonover at GE Aerospace, and Matt Robinson at Southwest Research Institute for their helpful discussion of the project and Terry Zhang for his help with the WAAM hardware.


\section*{Funding Data}

Research was sponsored by the ARM (Advanced Robotics for
Manufacturing) Institute through a grant from the Office of the
Secretary of Defense and was accomplished under Agreement
Number W911NF-17-3-0004. The views and conclusions con-
tained in this document are those of the authors and should
not be interpreted as representing the official policies, either
expressed or implied, of the Office of the Secretary of Defense
or the U.S. Government. The U.S. Government is authorized
to reproduce and distribute reprints for Government purposes
notwithstanding any copyright notation herein.


\bibliographystyle{asmejour}   

\bibliography{bib} 



\end{document}